\ifdefined\pdfoutput\pdfoutput=1\fi
\documentclass[11pt]{article}
\PassOptionsToPackage{table}{xcolor}
\providecommand{\aclmode}{preprint}
\usepackage[\aclmode]{acl}
\usepackage{times,latexsym}
\usepackage[T1]{fontenc}
\usepackage[utf8]{inputenc}
\usepackage{microtype}
\usepackage{graphicx}
\usepackage{amsmath,amssymb,amsthm}
\usepackage{booktabs,tabularx,array,longtable}
\usepackage{enumitem}
\usepackage{xcolor,url}
\usepackage{float,placeins}
\definecolor{inkblue}{RGB}{36,80,120}
\definecolor{rdred}{HTML}{AE2936}
\definecolor{sdblue}{HTML}{17518C}
\definecolor{plannerink}{HTML}{3D3846}
\definecolor{fixedgrey}{HTML}{7A7682}
\definecolor{singleink}{HTML}{1A181C}
\definecolor{accentorange}{HTML}{C2831F}
\definecolor{paleblue}{HTML}{A7CCE8}
\definecolor{tintred}{HTML}{F7E7E9}
\definecolor{tintblue}{HTML}{E6F0F7}
\definecolor{tintgrey}{HTML}{F4F3F5}
\definecolor{blueramp1}{HTML}{93B8D8}
\definecolor{blueramp2}{HTML}{6793BD}
\definecolor{blueramp3}{HTML}{3F70A2}
\definecolor{blueramp4}{HTML}{28588C}
\definecolor{blueramp5}{HTML}{113F6D}

\newcommand{\best}[1]{\textbf{\textcolor{rdred}{#1}}}
\newcommand{\second}[1]{\textcolor{sdblue}{\underline{#1}}}

\newcommand{\takeaway}[1]{#1}
\newcommand{\R}{\mathbb{R}}

\newcommand{\sg}{\operatorname{sg}}

\newcommand{\TimingGPU}{NVIDIA H800}
\newcommand{\LatencyRDThreeMs}{44.74}

\newcommand{\LatencyRDTwentyOneMs}{112.54}

\newcommand{\LatencyCheckpointGapPct}{1.4}
\newcommand{\LatencyNativeMs}{100.03}
\newcommand{\LatencyNativeCalls}{17.99}
\newcommand{\LatencyNativeInterpPct}{0.5}
\newcommand{\PrefillLongThreeS}{3.13}
\newcommand{\PrefillLongTwentyOneS}{3.23}
\newcommand{\PrefillLongIncreasePct}{3.2}
\newcommand{\CacheOneTwentyEightKiB}{364.1}
\newcommand{\CacheFiveTwelveKiB}{550.8}
\newcommand{\CacheTwentyFortyEightKiB}{1{,}297.8}
\newcommand{\PrefillRemeasuredNote}{}
\newcommand{\ContinuationSeedRange}{$0.0002$--$0.0024$}
\newcommand{\GroupedDeltaCI}{[+0.0002, +0.0014]}
\newcommand{\OracleCE}{2.9631}
\newcommand{\OracleStaticCE}{3.0203}
\newcommand{\OracleGain}{0.0572}

\newcommand{\RowsRecipeEarly}{%
3  & 4.2236 & \best{3.4108} \\
6  & 3.4551 & \best{3.3587} \\
11 & \best{3.3290} & 3.3490 \\
16 & \best{3.3114} & 3.3470 \\
21 & \best{3.3106} & 3.3468 \\
}

\newcommand{\RowsCost}{%
3 & (1,0,1,0,1,0) & 44.74 & 5.381 & 4.3787 & 82.2 & 3{,}134.6 \\
6 & (1,1,1,1,1,1) & 56.52 & 6.305 & 3.1941 & 94.3 & 3{,}145.6 \\
11 & (1,2,2,2,2,2) & 75.07 & 7.847 & 3.0465 & 112.2 & 3{,}171.6 \\
16 & (1,4,3,3,2,3) & 93.15 & 9.387 & 3.0316 & 128.4 & 3{,}194.5 \\
21 & (1,4,4,4,4,4) & 112.54 & 10.929 & 3.0307 & 147.3 & 3{,}233.7 \\
}

\newcommand{\RowsGroupedCache}{%
1{,}024 & 0.75 & 1.19 & 1.58$\times$ & 37\% \\
4{,}096 & 2.19 & 3.94 & 1.80$\times$ & 44\% \\
16{,}384 & 7.93 & 14.93 & 1.88$\times$ & 47\% \\
}
\newcommand{\RowsAllocDiag}{%
Static path & 6 & $-$0.05273 to $-$0.05053 \\
 & 11--21 & $-$0.00037 to +0.00060 \\
Static split & 6 & $-$0.05273 to $-$0.05053 \\
 & 11--21 & +0.00005 to +0.00063 \\
Mean totals & 21 & +0.00013 to +0.00014 \\
Shuffled totals & 21 & +0.00166 to +0.00167 \\
}
\newcommand{\RowsLongTwo}{%
4K & \second{25.483} & \best{24.934} & $-$0.0218 [$-$0.0252, $-$0.0184] \\
8K & \second{24.277} & \best{23.858} & $-$0.0174 [$-$0.0211, $-$0.0136] \\
16K & \second{23.986} & \best{23.660} & $-$0.0137 [$-$0.0176, $-$0.0097] \\
32K & \second{23.858} & \best{23.568} & $-$0.0122 [$-$0.0162, $-$0.0082] \\
64K & \second{23.806} & \best{23.644} & $-$0.0068 [$-$0.0110, $-$0.0025] \\
}
\newcommand{\RowsLongPlannerDiff}{%
4K & +3.9 [+0.6, +7.2] & +0.8 [+0.5, +1.3] \\
8K & +1.8 [$-$2.2, +5.4] & +1.3 [+0.9, +1.8] \\
16K & +1.9 [$-$1.8, +5.4] & +1.4 [+0.9, +1.8] \\
32K & +2.4 [$-$1.2, +5.8] & +1.3 [+0.9, +1.7] \\
64K & +6.4 [+2.8, +9.8] & +1.4 [+1.0, +1.8] \\
}
\newcommand{\RowsLongNiah}{%
4K & 61.67 & 68.33 & 68.33 & 68.33 \\
8K & 63.33 & 68.33 & 71.67 & 68.33 \\
16K & 66.67 & 70.00 & 70.00 & 70.00 \\
32K & 61.67 & 70.00 & 70.00 & 71.67 \\
64K & 53.33 & 63.33 & 65.00 & 58.33 \\
Overall & 61.33 & 68.00 & 69.00 & 67.33 \\
}

\newcommand{\RowsMainResults}{%
Single-pass (no loop) & 6 & \CacheOneTwentyEightKiB & 3.0506 & 20.1 & 30.4 & 62.6 & 50.7 & 51.6 & 25.4 & 40.15 & 180.4 & 43.20 \\
\midrule
LooM, Fixed4 & 21 & \CacheOneTwentyEightKiB & 3.0465 & 21.2 & 30.3 & 61.3 & 50.7 & 52.5 & 24.8 & 40.14 & 173.2 & 42.75 \\
LooM, Self-Distilled & 6 & \CacheOneTwentyEightKiB & 3.0640 & 20.8 & 30.5 & 61.8 & 51.5 & 51.0 & 24.3 & 39.98 & 168.6 & 43.60 \\
 & 21 & \CacheOneTwentyEightKiB & 3.0456 & 20.9 & 30.7 & 62.2 & 51.4 & 50.8 & 23.9 & 39.98 & 154.1 & 42.59 \\
LooM, Planner$^\dagger$ & 17.96 & \CacheOneTwentyEightKiB & 3.0355 & 21.4 & 31.0 & 61.6 & 51.9 & 53.0 & 24.1 & 40.52 & 153.9 & 42.63 \\
\midrule
\rowcolor{tintred}LooM, Random-Depth & 6 & \CacheOneTwentyEightKiB & 3.1941 & 15.8 & 30.5 & 62.6 & 51.4 & 51.8 & 25.3 & 39.57 & 292.6 & 52.47 \\
\rowcolor{tintred} & 11 & \CacheOneTwentyEightKiB & 3.0465 & 21.3 & 31.1 & 62.8 & 51.5 & 52.8 & 25.6 & 40.86 & 148.3 & 43.03 \\
\rowcolor{tintred} & 16 & \CacheOneTwentyEightKiB & \second{3.0316} & 21.6 & 31.2 & 62.8 & 52.3 & 53.1 & 25.4 & \best{41.09} & \second{138.2} & \second{42.16} \\
\rowcolor{tintred} & 21 & \CacheOneTwentyEightKiB & \best{3.0307} & 21.6 & 31.2 & 62.7 & 52.2 & 52.7 & 25.2 & \second{40.94} & \best{134.4} & \best{42.04} \\
}

\newcommand{\papertitle}{LooM: Loop-Invariant Memory for\texorpdfstring{\\}{ }Causal Encoder--Decoder Language Models}
\title{\papertitle}
\providecommand{\paperauthor}{%
{\mdseries Lele Yang$^{*}$\quad Muxi Diao$^{*}$\quad Zhonghao Yan\quad Jiankang Chen$^{\dagger}$\quad Bin Wen$^{\dagger}$}\\
Tingting Gao\quad Wentao Zhang\quad Kongming Liang\quad Zhanyu Ma\quad Weiran Xu$^{\ddagger}$\\[0.4em]
{\normalsize Beijing University of Posts and Telecommunications\quad Zhongguancun Academy}\\
{\normalsize Peking University\quad Kling Team}\\[0.2em]
{\normalsize\texttt{\{yang\_happy,\,dmx\}@bupt.edu.cn}}}
\providecommand{\pdfauthorname}{Lele Yang, Muxi Diao, Zhonghao Yan, Jiankang Chen, Bin Wen, Tingting Gao, Wentao Zhang, Kongming Liang, Zhanyu Ma, Weiran Xu}
\author{\paperauthor}
\hypersetup{pdftitle={\papertitle},pdfauthor={\pdfauthorname},colorlinks=true,linkcolor=inkblue,citecolor=inkblue,urlcolor=inkblue}
\begin{document}
\maketitle
{\renewcommand\thefootnote{}\footnotetext{\hspace*{-1.8em}$^{*}$Equal contribution. $^{\dagger}$Project lead. $^{\ddagger}$Corresponding author.}}
\begin{abstract}
Looped Transformers increase effective depth by re-executing shared layers, but three limitations hinder their adoption: memory grows with the loop count, compute can only be adjusted in whole-block steps, and the loop count is hard to change after training. We propose LooM (Loop-invariant Memory), which addresses each in turn. In a causal encoder--decoder (CED), LooM loops only the decoder layers after the first, which read a global memory written once per token, and each of them publishes the key-value (KV) entry for a token only on its first visit, so the cache does not grow with the loop count. These layers loop independently, providing every budget from 3 to 21 layer executions per token, and one model trained on randomly sampled budgets covers all of them without a router. With enough loops, LooM outperforms a single-pass baseline trained on the same data on language modeling, zero-shot accuracy and long-context perplexity, while using the same cache size. Our analysis further shows that the training budget distribution decides what extra depth buys. These results suggest that a CED is a natural place to loop: memory is written once, and depth becomes a budget chosen at deployment.
\end{abstract}
\section{Introduction}
\label{sec:intro}
\begin{figure*}[t]
\centering
\includegraphics[width=\textwidth]{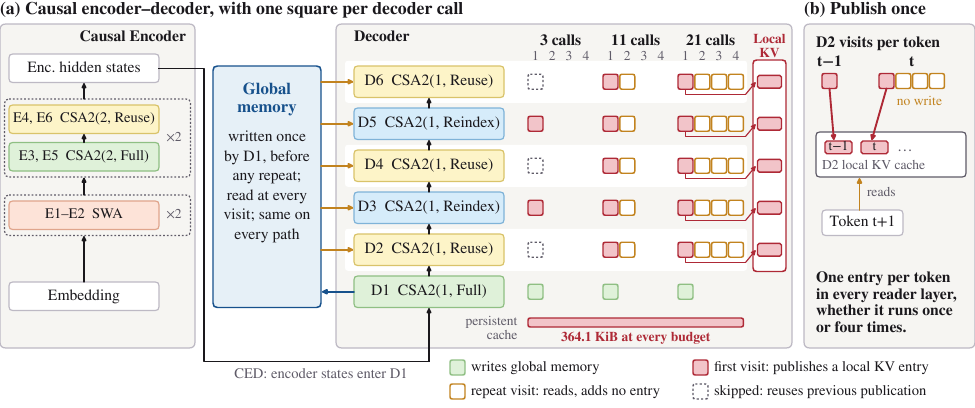}
\caption{Loop-invariant memory in LooM. (a) Decoder layer 1 (D1) writes global memory once per token, before any
repetition; each of D2--D6 then runs $d_\ell\in\{0,\ldots,4\}$ times (D3 and D5 at least once). Squares are
decoder calls on Random-Depth's locked paths at three budgets. Every visit re-reads global memory (left), but
only a layer's first visit publishes the token's local KV entry (right); publishing arrows are drawn for
the 21-call path only. (b) D2, standing for any of D2--D6: a later token reads one local entry per earlier
token however often the layer ran, so the cache (after a 128-token prompt) is the same on every path. CSA2
modes are defined in Section~\ref{sec:backbone}.}
\label{fig:teaser}
\end{figure*}
Looped Transformers apply the same stack of layers repeatedly to each token's hidden state, increasing effective depth without adding parameters \citep{universal,huginn,ouro,loopbeyondtwice,tflooped}. This added depth improves performance: looped models rival much larger standard models \citep{ouro} and scale test-time computation in latent space \citep{huginn}. However, three limitations hinder their wider adoption. (1)~\textbf{\textit{Memory grows with the loop count.}} In a looped Transformer, the same layers both build the context and refine each token's state, so every loop writes new key--value (KV) entries. With a standard KV cache, as in Ouro, memory therefore grows linearly with the number of loops \citep{ouro,melt}. (2)~\textbf{\textit{Compute can only be adjusted coarsely.}} Looped language models typically repeat a whole block of layers \citep{huginn,ouro,mor,asymloop}, so budgets come in multiples of that block and none lies below one full pass. (3)~\textbf{\textit{The loop count is hard to change after training.}} A model trained at one loop count degrades when run with fewer loops (Section~\ref{sec:frontier}). Varying compute therefore usually requires a learned router or halting policy \citep{act,ponder,mod,mor,ouro} or loop conditioning \citep{loopformer}.

To address these limitations, we propose LooM (Loop-invariant Memory), with one design for each. (1)~In a causal encoder--decoder (CED) \citep{ced}, the encoder and the first decoder layer write the global memory once per token, and LooM loops only the remaining decoder layers, which we call reader layers. Each reader layer writes (publishes) a token's local KV entry only on its first visit, so the cache does not grow with the loop count (Figure~\ref{fig:teaser}). (2)~Each reader layer loops independently, zero to four times, so one model offers every budget from 3 to 21 decoder calls (layer executions) per token; publish-once keeps the cache unchanged even when a layer is skipped. (3)~During training, each update samples a budget and then a random path, that is, how many times each reader layer runs, with that many calls, so the model learns to run at every budget. At inference we fix one path per budget, chosen on validation data, and need no router.

Trained on 5B tokens, the resulting 204M-parameter model beats a single-pass model trained on the same data from 11 calls on. At 21 calls its test cross-entropy is 0.020 nats lower and its zero-shot average 0.79 points higher at the same cache size. (Table~\ref{tab:main}).

Our contributions are:
\begin{itemize}[nosep,leftmargin=*]
\item \textbf{Loop-invariant memory.} We loop only the reader layers of a CED and publish each local KV entry once, so the cache does not grow with the loop count, even when layers loop independently or are skipped (Section~\ref{sec:cache}).
\item \textbf{Layer-level loops in one router-free model.} Each reader layer loops independently, giving budgets from 3 to 21 calls in steps of one layer execution; with Random-Depth training, one checkpoint serves them all (Section~\ref{sec:frontier}).
\item \textbf{An analysis of how to spend the budget.} The training budget distribution decides what extra depth buys (Section~\ref{sec:recipes}), repeated reads re-retrieve memory (Section~\ref{sec:reads}), and static paths nearly match a learned router (Section~\ref{sec:randomdepth}).
\end{itemize}
\section{Related Work}
\label{sec:related}
\paragraph{Looped models and their KV cache.}
Looped language models reuse shared layers to add depth without adding parameters \citep{huginn,ouro,loopbeyondtwice,tflooped,rrt,loopmoe}. Keeping a standard KV cache across iterations, as Ouro does, makes memory grow linearly with the loop count \citep{ouro,melt}. Mixture-of-Recursions can reuse the first recursion's KV \citep{mor}, MELT updates one cache per layer through a learned gate \citep{melt}, and concurrent work trains a looped stack to tolerate a shared cache or compresses the per-loop caches \citep{benice,loopedright2,lla}. These methods obtain a shared cache through training, fine-tuning or compression; in LooM it follows from where a CED writes memory, with no gate. The closest design, the concurrent Asymmetric Looped Transformer, also pairs a single-pass causal encoder with a weight-tied recurrent decoder \citep{asymloop}, but it loops the decoder as a unit for a fixed number of iterations and recomputes each local KV entry at every iteration, whereas our reader layers take independent depths and publish each entry once. We claim a CED execution rule, not the general idea of sharing KV across loops.

\paragraph{Reducing KV storage without loops.}
Cross-layer KV sharing saves storage across layers rather than loops \citep{yoco,lckv,cla}. Early-exit and layer-skipping methods must reconstruct the state of skipped layers \citep{calm,skipdecode,layerskip} or, as BUDDY does, leave the resulting cache misses open \citep{buddy}. Under publish-once, a skipped reader layer publishes nothing, and later tokens read the most recent preceding publication.

\paragraph{Setting a token's depth.}
Adaptive computation learns halting, exits or token routing \citep{act,universal,ponder,mod,mor,ouro,huginn,drllm}. BUDDY and LoopFormer serve several budgets from one checkpoint, through a per-step decision module or by conditioning each iteration on time and step size \citep{buddy,loopformer}. LooM needs neither at inference: like stochastic depth, LayerDrop and slimmable networks \citep{stochdepth,layerdrop,usnet}, Random-Depth samples each layer's depth during training, and a learned planner serves only as a comparison (Section~\ref{sec:randomdepth}).
\section{Looping the Read Stage}
\label{sec:method}
LooM loops only the decoder layers that read memory. This section follows one token through that loop, shows why repeated reads add no persistent memory, and covers training and inference.

\subsection{A Token's Path Through the CED}
\label{sec:backbone}
Our backbone adopts the causal encoder--decoder structure of DeepSeek-V4.1-Flash \citep{ced}, using dense FFNs. It also keeps that model's mHC and CSA2 components, and each attention layer runs in one of CSA2's three modes \citep{ced}. \textbf{Full} computes the layer's own KV entries and Top-$K$ retrieval support; \textbf{Reindex} reuses the KV but recomputes the support; \textbf{Reuse} takes both from the latest layer that computed them. Among the six encoder layers, layers 1--2 use sliding-window attention, layers 3 and 5 run in Full mode, and layers 4 and 6 run in Reuse mode. Among the six decoder layers, layer 1 runs in Full mode, layers 3 and 5 run in Reindex mode, and layers 2, 4, and 6 run in Reuse mode. Figure~\ref{fig:teaser} labels every block with its mode, and Appendix~\ref{app:training} gives the remaining architecture details.

Decoder memory has two forms. Global attention reads an uncompressed KV bank and indexer keys, both built by the encoder and the first decoder layer from the final encoder representation. Local attention reads a causal sliding-window history of width 128, which each decoder layer keeps for itself. Once the global bank is built, decoder layers 2--6 act as reader layers: at every visit they attend to the same bank and update their hidden states. Layers 3 and 5 also reselect Top-128 entries at each visit but add no global entries.

A token's path is $d=(1,d_2,\ldots,d_6)$. Decoder layer 1 runs once, and each of decoder layers 2--6 runs zero to four times in a row. Decoder layers 3 and 5 must run at least once because their Reindex mode has to produce new Top-$K$ support. These constraints give 2,000 legal paths. We loop each reader layer separately, rather than the decoder as a unit, for three reasons. Looping the unit allows only four budgets (6, 11, 16 and 21 calls), whereas per-layer depths reach every budget from 3 to 21. A depth of zero skips a layer, which gives the lowest budgets, such as the 3-call path $(1,0,1,0,1,0)$. Within a budget, calls can also be split unevenly across layers. All paths share weights and use no intermediate language-model head, no extra embedding, and no gradient detachment at recurrence boundaries. The $r$-th execution is
\begin{equation}
 s_{\ell}^{(r)}=F_\ell\big(s_{\ell}^{(r-1)};M_{\leq t},H_{\leq t,\ell},I_{\ell}^{(r)}\big),
\label{eq:reader}
\end{equation}
where $s_{\ell}^{(0)}$ is the preceding layer's output, $M_{\leq t}$ is the global memory already completed through token $t$, $H_{\leq t,\ell}$ is local history, and $I_{\ell}^{(r)}$ is current retrieval support. One decoder call is one layer execution for one token:
\begin{equation}
 C(d)=1+\textstyle\sum_{\ell=2}^{6}d_\ell\in[3,21].
\label{eq:cost}
\end{equation}
A budget is an allowed number of calls: decoder layer 1's single execution plus the loop counts of decoder layers 2--6. A static path is the layer-depth tuple used by every token at that budget. A cap is a per-token upper bound on calls, and the realized count can be lower. A suffix consists of decoder layers 2--6. It follows the prefix shared by every path: the encoder and decoder layer 1.

\subsection{Publish Once: What a Later Token Sees}
\label{sec:cache}
Publish-once gives the two memory pathways separate write rules. Global memory is complete before any decoder layer starts looping, so no loop ever writes to it. Local memory holds one KV entry per token in each decoder layer, and each execution of a layer on a token is a visit.

Consider a token $t$ with $d_2=4$, so that decoder layer 2 visits $t$ four times, and assume that its global memory $M_{\leq t}$ is complete before the first visit. All four visits update the hidden state of $t$. The first visit also publishes the token's local KV entry, that is, writes it to the cache where later tokens can read it. The other three visits recompute $t$'s query, which can shift its attention over the stored keys and values, but write no new entry. When token $t+1$ later reads its history at decoder layer 2, it therefore finds one local KV entry for token $t$, whether decoder layer 2 visited $t$ once or four times. Looping changes the hidden state used by the current token's computation and leaves unchanged the number of entries that later tokens can read. A skipped decoder layer publishes no entry of its own and reuses the most recent preceding publication; an indexing layer may change its retrieval support at each visit without writing new entries to global memory.

\paragraph{Cache size.} After a prefix of $T$ tokens, the persistent cache holds two parts: the global banks, written once per token by the encoder and decoder layer 1, and a sliding window of at most $W=128$ local entries in each of the 12 layers. Ignoring data type and metadata overhead, its size is
\begin{equation}
 \underbrace{\textstyle\sum_j n_j(T)\,k_j}_{\text{global banks}}+\underbrace{\textstyle\sum_{\ell=1}^{12}\min(T,W)\,k^{\mathrm{loc}}_\ell}_{\text{local windows}},
\label{eq:cachebound}
\end{equation}
where bank $j$ stores $n_j(T)$ positions of $k_j$ bytes and layer $\ell$ stores $k^{\mathrm{loc}}_\ell$ bytes per local entry. Neither term depends on the path: global memory is complete before any loop, and each layer publishes one local entry per token however often it runs. A 3-call and a 21-call path therefore store the same number of bytes. The stored values depend on the execution path, since a different path changes what later layers publish, but the cache layout and the bound do not.

A single cache layout therefore serves every compute budget. Rebuilding memory after looped reading would instead require running the looped decoder layers over every prompt position (Appendices~\ref{app:cost} and~\ref{app:grouped}).

\subsection{Training One Model for Every Budget}
\label{sec:training}
\looseness=-1 Publish-once lets the loop count vary without changing the cache layout, so we train a single model across loop counts with \textbf{Random-Depth}, and compare it against \textbf{Self-Distilled}, a control that adds a full-depth teacher. Both sample depths independently of token content, unroll each update to its largest sampled depth, and mask the unexecuted repetitions. Losses and schedules for both recipes are in Appendix~\ref{app:training}.

\textbf{Random-Depth} samples a budget and then a path of that budget, and it neither trains nor executes a router. It draws
\begin{equation}
 B\sim\pi,\qquad d\sim\mathrm{Unif}\{d:C(d)=B\},
\label{eq:randomdepth}
\end{equation}
where $\pi$ puts $1/16$ of its mass on budgets 3--10, $3/16$ on 11--16 and $3/4$ on 17--21, uniformly within each interval. The schedule is fixed for the whole run and gives 17.19 expected calls per token: three quarters of sampled paths fall in the 17--21 interval against 6.25\% at ten calls or fewer.

\textbf{Self-Distilled} samples a legal budget and path, runs that suffix alongside the full-depth suffix, and adds a stop-gradient Kullback--Leibler (KL) divergence from the full-depth distribution.

\subsection{Deployment with One Static Path per Budget}
\label{sec:static}
After training, inference needs one path for each number of calls. For each budget $B\in\{3,\ldots,21\}$, we select one static path on validation data, lock it before testing, and use it for all tokens.

The search runs in three stages, on 16, 128 and 512 validation samples. The first stage scores all 2,000 paths and keeps four candidates per exact cost, and the next two stages narrow these to two candidates and then one. Because only the first stage is exhaustive, a locked path is the best found, not a guaranteed optimum (Appendix~\ref{app:training}).

\subsection{A Learned Planner for Comparison}
\label{sec:planner}
To test whether token-dependent allocation helps, we ablate Random-Depth (Section~\ref{sec:randomdepth}) by replacing its static execution path with a learned planner. The planner reads decoder layer 1's normalized output $u_t$ and emits one categorical score group per reader layer,
\begin{equation}
 z_t=Au_t+b,
\label{eq:plannerscore}
\end{equation}
so a token's depths are fixed once, after the prefix is built and before any repetition.

The planner is trained with a straight-through estimator. For hard choice $a$ and relaxed probabilities $p$, candidate depth $r$ carries the weight $w_r=\mathbf{1}[r=a]+p_r-\sg(p_r)$, so the forward pass runs a single path while the backward pass compares the candidates at that layer. The training objective is
\begin{equation}
 \mathcal L=\mathcal L_{\mathrm{LM}}+\mathcal L_{\mathrm{index}}-\beta\mathcal H,
\label{eq:plannerloss}
\end{equation}
where $\mathcal L_{\mathrm{index}}$ is the indexer KL the architecture already carries and $\mathcal H$ is the mean routing entropy. The entropy bonus starts at $\beta=0.005$ and, together with exploration noise, keeps the policy from hardening too early; both anneal over the final tenth of training.

Eq.~\eqref{eq:plannerloss} rewards only language-modeling quality, so training teaches the planner where to spend depth, and the compute budget is imposed at inference. There, an exact dynamic program selects the highest-scoring legal path under a per-token cap; without a cap, we call the planner's own per-token choice native routing.
\section{Experimental Setup}
\label{sec:setup}
\begin{table}[t]
\centering\small
\setlength{\tabcolsep}{4pt}
\begin{tabularx}{\columnwidth}{@{}lX@{}}
\toprule
Model & Decoder depth during training \\
\midrule
Single-pass & Always 6 calls; no loop \\
\midrule
LooM, Fixed4 & Always 21 calls: decoder layers 2--6 run four times each \\
LooM, Random-Depth & A deep-heavy random budget (75\% at 17--21 calls), then a random path \\
LooM, Self-Distilled & A random path learns from a 21-call teacher \\
LooM, Planner & A learned router sets each layer's depth from the token \\
\midrule
Grouped memory & As Fixed4, but memory rebuilt at decoder layers 1, 3 and 5 \\
\bottomrule
\end{tabularx}
\caption{Models and controls. LooM models share the looped read stage and publish-once cache;
grouped memory writes memory inside the loop. Only the planner routes on token content.}
\label{tab:modelroles}
\end{table}
\begin{table*}[t]
\centering\footnotesize
\setlength{\tabcolsep}{2.5pt}
\begin{tabular}{lrrrrrrrrrrrr}
\toprule
& & Cache & Test & \multicolumn{6}{c}{Zero-shot accuracy (\%) $\uparrow$} & \multicolumn{1}{c}{} & \multicolumn{2}{c}{PPL $\downarrow$} \\
\cmidrule(lr){5-10}\cmidrule(lr){12-13}
Model & Calls & (KiB) & CE $\downarrow$ & LAMB. & HellaS. & PIQA & WinoG. & ARC-E & ARC-C & \multicolumn{1}{c}{\textbf{Avg}} & LAMB. & Wiki-2 \\
\midrule
\RowsMainResults
\bottomrule
\end{tabular}
\caption{Main results. All models are 204M-parameter CEDs trained on the same data and updates; LooM rows
differ only in training recipe and use one validation-locked static path per budget ($^\dagger$native routing,
17.96 calls on average). Cache is measured after a 128-token prompt; CE is in nats per token. Zero-shot scores are
accuracy (\%), normalized for HellaSwag and ARC-C. As several tasks are near chance, only CE, Avg and PPL are
marked (bold red: best; underlined blue: second). CE at all budgets: Figure~\ref{fig:frontier}.}
\label{tab:main}
\end{table*}
\paragraph{Models and controls.}
Table~\ref{tab:modelroles} lists every run. All looped models except grouped memory are LooM models: they share the looped read stage and the publish-once cache of Section~\ref{sec:method} and differ only in how decoder depth is set during training. Random-Depth is our main multi-budget model (Section~\ref{sec:training}). Each remaining run isolates one factor against it. Single-pass and Fixed4 hold the decoder at 6 and 21 calls, separating the effect of looping from the effect of depth sampling; Self-Distilled changes which paths are trained, so it tests how the training distribution shapes low budgets; the planner (Section~\ref{sec:planner}) replaces the static path with a learned router, so it measures what token-dependent allocation adds; and grouped memory keeps Fixed4's schedule but writes context memory at several decoder layers, so it isolates where memory is written.

\paragraph{Training.}
All runs share a 204M-parameter backbone and are trained from scratch on 5B valid target tokens of FineWeb-Edu \citep{finewebedu} with the same seed, batching and optimizer schedule, so every comparison is matched in parameters and training data (Appendix~\ref{app:data}; configuration in Table~\ref{tab:architecture}). They are not matched in training compute: a looped update runs more decoder calls than a single-pass one, and the looped runs used 2.9--4.7$\times$ the single-pass run's GPU-hours.

\paragraph{Language modeling.}
We report token-weighted test CE on 2,048 held-out sequences, with paired document-bootstrap intervals over 7,836 segments. Perplexity (PPL) is $\exp(\mathrm{CE})$.

\paragraph{Downstream and long-context evaluation.}
The pre-specified standard suite is LAMBADA \citep{lambada}, HellaSwag \citep{hellaswag}, PIQA \citep{piqa}, WinoGrande \citep{winogrande}, ARC-Easy and ARC-Challenge \citep{arc}, plus WikiText-2 word perplexity \citep{wikitext}; we fixed the metrics and the macro average before scoring (Appendix~\ref{app:downstream}). Scoring uses lm-evaluation-harness 0.4.13 \citep{lmeval}. After a shared continuation to 64K context, Random-Depth, the single-pass model and the planner are evaluated on PG19 \citep{pg19} and three RULER retrieval tasks \citep{ruler} (Appendix~\ref{app:long}).

\paragraph{Cost.}
We report single-sequence decoding latency (batch size one), where the per-token cost of extra calls is most visible, measured with a reference implementation on one \TimingGPU{} GPU. Matrix FLOPs count linear projections, attention $QK^\top$ and $AV$, indexer/router projections and vocabulary head operations from operand shapes, including tile padding and masked bank rows; they exclude normalization, softmax, sorting, indexing, cache copies and pointwise reductions. For Fixed4's 21-call latency we reuse Random-Depth's path shape; on identical paths, checkpoints differ by at most \LatencyCheckpointGapPct\% (Appendix~\ref{app:cost}). Following DeepSeek-V4.1-Flash \citep{ced}, prefill uses sliding-window attention (SWA) replay over the most recent 128 prompt tokens, which approximately reconstructs the decoder's SWA KV cache without running the decoder over the prompt. As a result, 65K-token prefill grows only \PrefillLongIncreasePct\% from 3 to 21 calls, so the budget mainly affects decode.
\section{Results and Analysis}
\label{sec:results}
This section first reads the main results (Table~\ref{tab:main}): quality, cache size and downstream accuracy. It then examines the role of the training depth distribution, the compute cost of additional loops, what the repeated reads contribute, whether tokens need different depths, and long-context evaluation after 64K continuation. Except where we state native routing, every model uses one validation-locked static path per budget.

\subsection{Main Results: One Checkpoint across Budgets}
\label{sec:frontier}
We first ask whether LooM, by repeating the read stage, improves on one ordinary forward pass, and whether a single checkpoint can do so across loop counts. Table~\ref{tab:main} gathers the answer. Two controls frame it: the single-pass model shows what the same backbone reaches at six calls, and Fixed4 is trained only on the full 21-call loop and is evaluated on shortened paths.
\begin{figure}[t]
\centering
\includegraphics[width=\columnwidth]{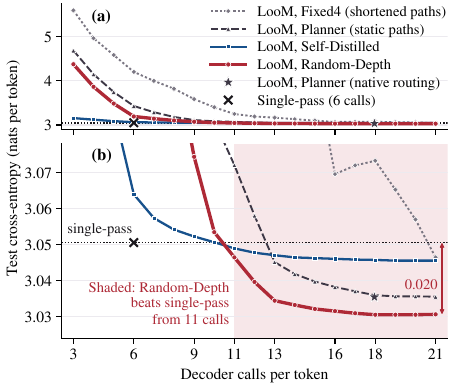}
\caption{Test CE (nats per token; lower is better) of LooM models on validation-locked static paths; the
star is the planner's native routing. (a) Full range. (b) Zoom: Random-Depth and Self-Distilled both cross
the six-call single-pass line between 10 and 11 calls.}
\label{fig:frontier}
\end{figure}
\takeaway{One LooM checkpoint, trained with Random-Depth, covers 3--21 calls and beats the six-call single-pass model starting at 
11 calls, by 0.020 nats at 21.}

\paragraph{Quality.} The crossing lies between 10 and 11 calls (Figure~\ref{fig:frontier}b), and the 0.020-nat gain at 21 calls is a 2.0\% reduction in perplexity. CE falls by 0.0158 nats from 11 to 21 calls and is nearly flat from 16 calls on, so most of the gain from extra budget arrives early. Below 11 calls both multi-budget recipes trail the single-pass model, which is trained for exactly six. Self-Distilled is the stronger recipe at that low end, at the cost of a full-depth teacher during training, and Random-Depth from 11 calls on.

\paragraph{One checkpoint versus a fixed loop.} Fixed4 shows what training across loop counts adds. Trained only on its 21-call loop, it degrades when shortened (Figure~\ref{fig:frontier}a), and on that same path Random-Depth is 0.0158 nats better ($[-0.0164,-0.0151]$). Random-Depth at 11 calls already matches Fixed4 at 21: both reach 3.0465, a difference of $+0.00003$ ($[-0.00065,+0.00071]$).

\paragraph{Cache.} Every row of Table~\ref{tab:main} stores the same \CacheOneTwentyEightKiB\,KiB after a 128-token prompt, across models and across Random-Depth's budgets, as Eq.~\eqref{eq:cachebound} predicts. Section~\ref{sec:quality} covers their compute cost.

\paragraph{Downstream accuracy.} The zero-shot columns follow the CE frontier. At 21 calls Random-Depth improves the six-task average over the single-pass model by 0.79 points ($[+0.05,+1.54]$) and lowers LAMBADA perplexity from 180.4 to 134.4. At 6 calls it trails the single-pass model; at 11 calls it is 0.71 points above Fixed4 at 21 ($[-0.04,+1.43]$); and the average peaks at 16 calls. Several tasks are near chance at this scale, so we read only the average and the perplexities.

\subsection{The Training Distribution Decides What Extra Depth Buys}
\label{sec:recipes}
\takeaway{Spreading the training samples evenly across budgets yields a model whose quality barely responds
to depth. Random-Depth concentrates them on the deployable range and gains twelve times more from the same
extra calls.}

Random-Depth's schedule is deep-heavy (Section~\ref{sec:training}). Its quality is likewise strong at high budgets and weak at low ones (Figure~\ref{fig:frontier}). The correspondence alone does not establish cause: weak low-budget paths could reflect a cost intrinsic to shallow execution rather than too few shallow training samples. Self-Distilled and a uniform-budget control separate the two by changing the trained paths (Appendix~\ref{app:uniform}).

Shallow execution carries no intrinsic penalty. At three calls Self-Distilled reaches 3.1570 where Random-Depth reaches 4.3787 (Figure~\ref{fig:frontier}a), on the same architecture, data and updates; a uniform-budget control shows the same effect (Table~\ref{tab:recipes}). The shortfall at low budgets therefore follows from the training distribution.

What a flat distribution costs is the response to depth. Early in training, the uniform-budget control improves by only 0.012 nats from 6 to 21 calls, against 0.145 for Random-Depth's deep-heavy schedule---twelve times less return on the same fifteen extra calls (Table~\ref{tab:recipes}); a model whose quality does not respond to depth cannot serve a compute budget at all. Random-Depth is best at 16 and 21 calls here, and after full training it leads from 11 calls on---the only region in which any recipe overtakes the single-pass baseline (Section~\ref{sec:frontier}). Self-Distilled stays within 0.015 nats of the best recipe at every budget, but it runs a full-depth teacher at every update, at 4.73$\times$ the single-pass run's GPU-hours against Random-Depth's 2.94$\times$, and its advantage does not reach the downstream suite (Appendix~\ref{app:downstream}).

\subsection{Extra Loops Cost Compute, Not Persistent Memory}
\label{sec:quality}
\takeaway{Extra loops increase compute while leaving cache size unchanged; rebuilding memory inside the loop enlarges the cache without improving quality.}

The cache is independent of the loop count at longer prompts too: after 512- and 2,048-token prompts it is \CacheFiveTwelveKiB{} and \CacheTwentyFortyEightKiB\,KiB, respectively, on both the 3- and 21-call paths.

\begin{table*}[t]
\centering\small
\setlength{\tabcolsep}{4pt}
\begin{tabular}{rlrrrrr}
\toprule
& & \multicolumn{3}{c}{Decode, 128-token prompt} & \multicolumn{2}{c}{Prefill (ms)} \\
\cmidrule(lr){3-5}\cmidrule(lr){6-7}
Calls & Path & ms/token & GFLOP/token & Test CE $\downarrow$ & 4,096 & 65,536 \\
\midrule
\RowsCost
\bottomrule
\end{tabular}
\caption{Cost of Random-Depth's locked paths at batch size one on one \TimingGPU{} GPU with a Python
reference implementation (medians of 24 decode and five prefill runs). Prefill columns use 4,096- and
65,536-token prompts. Full measurements are in Appendix~\ref{app:cost}.}
\label{tab:cost}
\end{table*}

\looseness=-1 The cost of extra loops therefore falls on compute. At batch size one, reference decode latency grows from \LatencyRDThreeMs\,ms at 3 calls to \LatencyRDTwentyOneMs\,ms at 21 calls, whereas 65K-token prefill grows by only \PrefillLongIncreasePct\% (Table~\ref{tab:cost}; Appendix~\ref{app:cost}). Relative to the six-call single-pass model, Random-Depth's 21-call gain requires 99.1\% higher reference decode latency (56.52 to 112.54 ms) and 73.3\% more matrix FLOPs (6.305 to 10.929 GFLOP per token). Because Random-Depth reaches Fixed4's 21-call quality at 11 calls, this smaller budget reduces latency by 33.3\% and matrix FLOPs by 28.2\% (Table~\ref{tab:main}; Appendix~\ref{app:cost}). All of these measurements come from a reference implementation without optimized kernels.

\looseness=-1 Since LooM keeps the cache fixed, the remaining question is whether rebuilding memory inside the loop improves quality enough to justify its extra storage. The grouped-memory variant tests this by rebuilding decoder memory at decoder layers 1, 3 and 5 under Fixed4's schedule. It increases cache size by 1.58--1.88$\times$ over 1,024--16,384-token prompts and is slightly worse in test CE than LooM on the same schedule (3.04728 against 3.04646, a difference of $+0.00082$, \GroupedDeltaCI; Table~\ref{tab:grouped}; Appendix~\ref{app:grouped}).

\begin{table}[t]
\centering\small
\setlength{\tabcolsep}{4pt}
\begin{tabular}{rrrrr}
\toprule
Prompt & LooM & Grouped & Ratio & LooM saves \\
\midrule
\RowsGroupedCache
\bottomrule
\end{tabular}
\caption{Persistent decode cache (MiB, BF16) of LooM and grouped memory under Fixed4's
21-call schedule.}
\label{tab:grouped}
\end{table}

\subsection{What Repeated Reads Do}
\label{sec:reads}
\takeaway{Repeated reads both refine the hidden state and re-retrieve memory: freezing the first visit's retrieval selections costs 0.0114 nats.}

Re-executing a decoder layer can change a token's computation in two ways. Each visit updates the token's hidden state, and each visit selects which memory entries to read. We measure these two effects separately.

Repeated reads refine the hidden state. The first visit of each layer makes the largest update, and later visits make smaller ones, in Random-Depth and the planner alike (Figure~\ref{fig:mechanism}b).

Repeated reads also re-retrieve memory. For decoder layer 3, the first- and fourth-visit Top-$K$ sets have Jaccard similarity 0.54 overall and 0.30 beyond position 2,048, and 32\% of the fourth visit's global attention falls on entries the first visit did not select. To test whether this re-retrieval matters for quality, we freeze each layer's first-visit selections and reuse them on all later visits. This intervention costs Random-Depth 0.0114 nats ($[0.0106,0.0122]$), so later visits gain quality from the entries they newly select.

A repeated read therefore does two things at once: it applies a progressively smaller refinement to the token's state, and it retrieves memory entries that the first visit did not select.

\subsection{Token-Specific Depth Allocation}
\label{sec:randomdepth}
\takeaway{Do tokens need different depths? At 4K, one static path per budget nearly matches the tested linear
planner; after 64K continuation, native routing has a small advantage.}

\paragraph{Deployable routing at 4K.}
\looseness=-1 Compared within one checkpoint, the planner's native routing at 17.96 calls is only 0.00051 nats
better than its own static 18-call path. Table~\ref{tab:allocdiagmain} reports four ways of replacing native
routing, on two calibrated planner continuations (Appendix~\ref{app:budget}):
\begin{itemize}[nosep,leftmargin=*]
\item \textbf{Static path} gives every token one validation-selected path per cap.
\item \textbf{Static split} keeps each token's native total but replaces its layer split with a
validation-selected path of equal total.
\item \textbf{Mean totals} keeps native splits but gives each token its document's mean total.
\item \textbf{Shuffled totals} permutes totals per document.
\end{itemize}

At caps 11--21, static path and static split change CE by less than $10^{-3}$. At cap 21, mean totals costs only 0.00013--0.00014 nats and shuffled totals costs 0.0017. The planner also orders equal-cost alternatives correctly only 51.4--51.5\% of the time. A coarser scalar-depth planner, which predicts one depth per token shared by every reader layer, trails both the layerwise planner and Random-Depth at a comparable budget (Appendix~\ref{app:scalar}). Besides token dependence, the planner's objective adds an entropy bonus and exploration noise to Random-Depth's (Section~\ref{sec:planner}). These results bound what a single linear planner of roughly 20K parameters contributes at 4K; they do not bound the value of adaptive computation in general.

\paragraph{Potential gains from allocation.}
\label{sec:mechanism}
A per-token allocation can help only if the benefit of extra decoder calls differs across tokens, so we first measure how this benefit is distributed. We use the gain from 11 to 21 calls, because a comparison that starts at six calls would mostly measure the shortfall of Random-Depth's weak six-call path. We group test tokens into deciles by difficulty, defined as the target negative log-likelihood (NLL), and the 11-to-21 gain rises from 0.002 to 0.025 nats across these deciles, about twelvefold (Figure~\ref{fig:mechanism}a; Appendix~\ref{app:mechanism}). The target NLL depends on the correct next token, which a deployed router cannot observe when it chooses a depth. The deciles therefore show that the benefit of extra calls varies with token difficulty, but they do not show that a router can identify the tokens that benefit most from the prefix alone.

To estimate how much per-token allocation could help, we apply a hindsight policy to the same 256 test sequences. For each token, the policy selects one of the locked 11-, 13-, 16- and 21-call paths according to their target NLLs, and the selections are adjusted so that the average cost is exactly 16 calls. We then execute the mixed paths, so that later tokens read the local-KV entries actually published by the chosen paths. The mixed paths reach a CE of \OracleCE{}, compared with \OracleStaticCE{} for the static 16-call path, a gain of \OracleGain{} nats (Appendix~\ref{app:oracle}). The step that enforces the 16-call average is not guaranteed to be optimal, so this gain is a lower bound on the best hindsight gain available from these four paths. It also does not bound the gain achievable by routers in general.

\paragraph{After 64K continuation.}
The continued planner with native routing has lower PG19 perplexity than Random-Depth, by 0.0039 nats at 4K and 0.0064 at 64K (Section~\ref{sec:long}). Its native routing improves over its own static 21-call path by only 0.0008--0.0014 nats, so most of that advantage does not come from per-token routing. The continued models therefore agree with the 4K finding that one static path per budget nearly matches the planner, and add a small native-routing advantage. All of these differences are below 0.01 nats on one training seed (Appendix~\ref{app:long}).

\subsection{Long-Context Evaluation}
\label{sec:long}
\label{sec:downstream}
The preceding comparisons use the 4K training context. Random-Depth, the single-pass model and the planner then receive the same 600-update continuation, including 360 updates at 64K.

\paragraph{Language modeling.}
Random-Depth keeps lower PG19 perplexity than the single-pass model at every evaluated length from 4K to 64K, and every interval excludes zero (Table~\ref{tab:longmain}). The gap narrows with length, from 0.0218 nats at 4K to 0.0068 at 64K.

\begin{table}[t]
\centering\footnotesize
\setlength{\tabcolsep}{4pt}
\begin{tabular}{lrrl}
\toprule
Ctx. & Single & RD & RD $-$ Single, 95\% CI \\
\midrule
\RowsLongTwo
\bottomrule
\end{tabular}
\caption{PG19 perplexity after the shared 64K continuation. The last column is Random-Depth (RD) minus
single-pass NLL in nats, with a whole-book 95\% CI; negative favors RD. Bold red: best; underlined blue: second
best at each length.}
\label{tab:longmain}
\end{table}

\paragraph{Retrieval.}
We apply identical answer-format fine-tuning to each model and evaluate on three needle-in-a-haystack tasks (Appendix~\ref{app:long}). Random-Depth reaches 67.33\% accuracy against 61.33\% for the single-pass model, a gain of 6.0 points with interval $[+1.33,+10.33]$. On these 300 questions, Random-Depth matches the planner.

\section{Conclusion}
\label{sec:conclusion}
We presented LooM, which loops only the reader layers of a causal encoder--decoder and publishes each local KV entry once, so the cache stays fixed however often a layer runs. Per-layer loops give every budget from 3 to 21 calls, and Random-Depth training lets one router-free checkpoint serve them all. With enough loops, LooM outperforms a single-pass model trained on the same data in language modeling, zero-shot accuracy and long-context evaluation, at the same cache size. Looping inside a CED thus turns depth into a compute budget chosen at deployment; scaling it to larger models is the natural next step.
\label{maintext:end}
\section*{Limitations}
Our evidence comes from one 204M-parameter backbone, 5B training tokens and one seed, so we cannot say how the gains move with scale; latency is likewise a batch-size-one reference measurement, not optimized serving throughput. The baselines are matched in parameters and data but not in compute: no non-looped model is given the FLOPs of a 21-call path, and the whole-decoder loop that keeps per-loop caches is a preliminary, partly trained run (Appendix~\ref{app:ouro}).
\bibliography{references}

\begin{thebibliography}{43}
\providecommand{\natexlab}[1]{#1}

\bibitem[{Bae et~al.(2025{\natexlab{a}})Bae, Fisch, Harutyunyan, Ji, Kim, and
  Schuster}]{rrt}
Sangmin Bae, Adam Fisch, Hrayr Harutyunyan, Ziwei Ji, Seungyeon Kim, and Tal
  Schuster. 2025{\natexlab{a}}.
\newblock \href {https://arxiv.org/abs/2410.20672} {{Relaxed Recursive
  Transformers: Effective Parameter Sharing with Layer-wise LoRA}}.
\newblock In \emph{International Conference on Learning Representations}.

\bibitem[{Bae et~al.(2025{\natexlab{b}})Bae, Kim, Bayat, Kim, Ha, Schuster,
  Fisch, Harutyunyan, Ji, Courville, and Yun}]{mor}
Sangmin Bae, Yujin Kim, Reza Bayat, Sungnyun Kim, Jiyoun Ha, Tal Schuster, Adam
  Fisch, Hrayr Harutyunyan, Ziwei Ji, Aaron Courville, and Se-Young Yun.
  2025{\natexlab{b}}.
\newblock \href {https://arxiv.org/abs/2507.10524} {{Mixture-of-Recursions:
  Learning Dynamic Recursive Depths for Adaptive Token-Level Computation}}.
\newblock \emph{arXiv preprint arXiv:2507.10524}.

\bibitem[{Banino et~al.(2021)Banino, Balaguer, and Blundell}]{ponder}
Andrea Banino, Jan Balaguer, and Charles Blundell. 2021.
\newblock \href {https://arxiv.org/abs/2107.05407} {{PonderNet: Learning to
  Ponder}}.
\newblock \emph{arXiv preprint arXiv:2107.05407}.

\bibitem[{Ben~Allal et~al.(2025)Ben~Allal, Lozhkov, Bakouch,
  Mart{\'i}n~Bl{\'a}zquez, Penedo, Tunstall, Marafioti, Kydl{\'i}{\v c}ek,
  Piqueres~Lajar{\'i}n, Srivastav, Lochner, Fahlgren, Nguyen, Fourrier,
  Burtenshaw, Larcher, Zhao, Zakka, Morlon, Raffel, von Werra, and
  Wolf}]{smollm2}
Loubna Ben~Allal, Anton Lozhkov, Elie Bakouch, Gabriel Mart{\'i}n~Bl{\'a}zquez,
  Guilherme Penedo, Lewis Tunstall, Andr{\'e}s Marafioti, Hynek Kydl{\'i}{\v
  c}ek, Agust{\'i}n Piqueres~Lajar{\'i}n, Vaibhav Srivastav, Joshua Lochner,
  Caleb Fahlgren, Xuan-Son Nguyen, Cl{\'e}mentine Fourrier, Ben Burtenshaw,
  Hugo Larcher, Haojun Zhao, Cyril Zakka, Mathieu Morlon, and 3 others. 2025.
\newblock \href {https://arxiv.org/abs/2502.02737} {{SmolLM2: When Smol Goes
  Big -- Data-Centric Training of a Small Language Model}}.
\newblock \emph{arXiv preprint arXiv:2502.02737}.

\bibitem[{Bisk et~al.(2020)Bisk, Zellers, Le~Bras, Gao, and Choi}]{piqa}
Yonatan Bisk, Rowan Zellers, Ronan Le~Bras, Jianfeng Gao, and Yejin Choi. 2020.
\newblock \href {https://ojs.aaai.org/index.php/AAAI/article/view/6239} {{PIQA:
  Reasoning about Physical Commonsense in Natural Language}}.
\newblock In \emph{Proceedings of the AAAI Conference on Artificial
  Intelligence}.

\bibitem[{Brandon et~al.(2024)Brandon, Mishra, Nrusimha, Panda, and
  Ragan-Kelley}]{cla}
William Brandon, Mayank Mishra, Aniruddha Nrusimha, Rameswar Panda, and
  Jonathan Ragan-Kelley. 2024.
\newblock \href {https://doi.org/10.52202/079017-2758} {{Reducing Transformer
  Key-Value Cache Size with Cross-Layer Attention}}.
\newblock In \emph{Advances in Neural Information Processing Systems}.

\bibitem[{Chen et~al.(2026{\natexlab{a}})Chen, Li, Liang, Lao, and
  Liu}]{tflooped}
Lizhang Chen, Jonathan Li, Chen Liang, Ni~Lao, and Qiang Liu.
  2026{\natexlab{a}}.
\newblock \href {https://arxiv.org/abs/2605.23872} {{Training-Free Looped
  Transformers}}.
\newblock \emph{arXiv preprint arXiv:2605.23872}.

\bibitem[{Chen et~al.(2026{\natexlab{b}})Chen, Li, Huang, Yin, Shang, and
  Qin}]{loopmoe}
Wenkai Chen, Tianshu Li, Wenyong Huang, Yichun Yin, Lifeng Shang, and Chengwei
  Qin. 2026{\natexlab{b}}.
\newblock \href {https://arxiv.org/abs/2606.04438} {{LoopMoE: Unifying
  Iterative Computation with Mixture-of-Experts for Language Modeling}}.
\newblock \emph{arXiv preprint arXiv:2606.04438}.

\bibitem[{Clark et~al.(2018)Clark, Cowhey, Etzioni, Khot, Sabharwal, Schoenick,
  and Tafjord}]{arc}
Peter Clark, Isaac Cowhey, Oren Etzioni, Tushar Khot, Ashish Sabharwal, Carissa
  Schoenick, and Oyvind Tafjord. 2018.
\newblock \href {https://arxiv.org/abs/1803.05457} {{Think you have Solved
  Question Answering? Try ARC, the AI2 Reasoning Challenge}}.
\newblock \emph{arXiv preprint arXiv:1803.05457}.

\bibitem[{Conchello~Vendrell et~al.(2026)Conchello~Vendrell,
  Padr{\'e}s~Masdemont, Grillo, Ros-Giralt, Behboodi, and Massoli}]{melt}
Victor Conchello~Vendrell, Arnau Padr{\'e}s~Masdemont, Niccol{\`o} Grillo,
  Jordi Ros-Giralt, Arash Behboodi, and Fabio~Valerio Massoli. 2026.
\newblock \href {https://arxiv.org/abs/2605.07721} {{Memory-Efficient Looped
  Transformer: Decoupling Compute from Memory in Looped Language Models}}.
\newblock \emph{arXiv preprint arXiv:2605.07721v3}.

\bibitem[{{DeepSeek-AI}(2026)}]{ced}
{DeepSeek-AI}. 2026.
\newblock \href {https://arxiv.org/abs/2609.19969} {{DeepSeek-V4.1-Flash:
  Pushing the Limits of KV Cache Compression}}.
\newblock \emph{arXiv preprint arXiv:2609.19969}.

\bibitem[{Dehghani et~al.(2019)Dehghani, Gouws, Vinyals, Uszkoreit, and
  Kaiser}]{universal}
Mostafa Dehghani, Stephan Gouws, Oriol Vinyals, Jakob Uszkoreit, and {\L}ukasz
  Kaiser. 2019.
\newblock \href {https://arxiv.org/abs/1807.03819} {{Universal Transformers}}.
\newblock In \emph{International Conference on Learning Representations}.

\bibitem[{Del~Corro et~al.(2023)Del~Corro, Del~Giorno, Agarwal, Yu, Awadallah,
  and Mukherjee}]{skipdecode}
Luciano Del~Corro, Allison Del~Giorno, Sahaj Agarwal, Bin Yu, Ahmed~Hassan
  Awadallah, and Subhabrata Mukherjee. 2023.
\newblock \href {https://arxiv.org/abs/2307.02628} {{SkipDecode: Autoregressive
  Skip Decoding with Batching and Caching for Efficient LLM Inference}}.
\newblock \emph{arXiv preprint arXiv:2307.02628}.

\bibitem[{Elhoushi et~al.(2024)Elhoushi, Shrivastava, Liskovich, Hosmer, Wasti,
  Lai, Mahmoud, Acun, Agarwal, Roman, Aly, Chen, and Wu}]{layerskip}
Mostafa Elhoushi, Akshat Shrivastava, Diana Liskovich, Basil Hosmer, Bram
  Wasti, Liangzhen Lai, Anas Mahmoud, Bilge Acun, Saurabh Agarwal, Ahmed Roman,
  Ahmed~A Aly, Beidi Chen, and Carole-Jean Wu. 2024.
\newblock \href {https://arxiv.org/abs/2404.16710} {{LayerSkip: Enabling Early
  Exit Inference and Self-Speculative Decoding}}.
\newblock In \emph{Proceedings of the 62nd Annual Meeting of the Association
  for Computational Linguistics}.

\bibitem[{Fan et~al.(2020)Fan, Grave, and Joulin}]{layerdrop}
Angela Fan, Edouard Grave, and Armand Joulin. 2020.
\newblock \href {https://arxiv.org/abs/1909.11556} {{Reducing Transformer Depth
  on Demand with Structured Dropout}}.
\newblock In \emph{International Conference on Learning Representations}.

\bibitem[{Gao et~al.(2023)Gao, Tow, Abbasi, Biderman, Black, DiPofi, Foster,
  Golding, Hsu, Le~Noac'h, Li, McDonell, Muennighoff, Ociepa, Phang, Reynolds,
  Schoelkopf, Skowron, Sutawika, Tang, Thite, Wang, Wang, and Zou}]{lmeval}
Leo Gao, Jonathan Tow, Baber Abbasi, Stella Biderman, Sid Black, Anthony
  DiPofi, Charles Foster, Laurence Golding, Jeffrey Hsu, Alain Le~Noac'h,
  Haonan Li, Kyle McDonell, Niklas Muennighoff, Chris Ociepa, Jason Phang,
  Laria Reynolds, Hailey Schoelkopf, Aviya Skowron, Lintang Sutawika, and 5
  others. 2023.
\newblock \href {https://doi.org/10.5281/zenodo.10256836} {{A Framework for
  Few-Shot Language Model Evaluation}}.

\bibitem[{Geiping et~al.(2025)Geiping, McLeish, Jain, Kirchenbauer, Singh,
  Bartoldson, Kailkhura, Bhatele, and Goldstein}]{huginn}
Jonas Geiping, Sean McLeish, Neel Jain, John Kirchenbauer, Siddharth Singh,
  Brian~R. Bartoldson, Bhavya Kailkhura, Abhinav Bhatele, and Tom Goldstein.
  2025.
\newblock \href {https://arxiv.org/abs/2502.05171} {{Scaling up Test-Time
  Compute with Latent Reasoning: A Recurrent Depth Approach}}.
\newblock \emph{arXiv preprint arXiv:2502.05171}.

\bibitem[{Graves(2016)}]{act}
Alex Graves. 2016.
\newblock \href {https://arxiv.org/abs/1603.08983} {{Adaptive Computation Time
  for Recurrent Neural Networks}}.
\newblock \emph{arXiv preprint arXiv:1603.08983}.

\bibitem[{He et~al.(2026)He, Li, Chang, Meng, Yin, and Liu}]{loopbeyondtwice}
Di~He, Pengxiang Li, Da~Chang, Qingyan Meng, Lu~Yin, and Shiwei Liu. 2026.
\newblock \href {https://arxiv.org/abs/2610.01153} {{Looping Beyond Twice: A
  Scalable Recipe for Looped Mixture-of-Experts}}.
\newblock \emph{arXiv preprint arXiv:2610.01153}.

\bibitem[{Heakl et~al.(2025)Heakl, Gubri, Khan, Yun, and Oh}]{drllm}
Ahmed Heakl, Martin Gubri, Salman Khan, Sangdoo Yun, and Seong~Joon Oh. 2025.
\newblock \href {https://arxiv.org/abs/2510.12773} {{Dr.LLM: Dynamic Layer
  Routing in LLMs}}.
\newblock \emph{arXiv preprint arXiv:2510.12773}.

\bibitem[{Hsieh et~al.(2024)Hsieh, Sun, Kriman, Acharya, Rekesh, Jia, Zhang,
  and Ginsburg}]{ruler}
Cheng-Ping Hsieh, Simeng Sun, Samuel Kriman, Shantanu Acharya, Dima Rekesh, Fei
  Jia, Yang Zhang, and Boris Ginsburg. 2024.
\newblock \href {https://arxiv.org/abs/2404.06654} {{RULER: What's the Real
  Context Size of Your Long-Context Language Models?}}
\newblock In \emph{First Conference on Language Modeling}.

\bibitem[{Huang et~al.(2026)Huang, Shi, Chen, Wen, Liu, Xing, and
  Ma}]{loopedright2}
Benhao Huang, Chufan Shi, Junlin Chen, Shicheng Wen, Zhengzhong Liu, Eric Xing,
  and Xuezhe Ma. 2026.
\newblock \href {https://arxiv.org/abs/2610.06833} {{Towards Looped Models Done
  Right, Part II: Rethinking at Fixed Points}}.
\newblock \emph{arXiv preprint arXiv:2610.06833}.

\bibitem[{Huang et~al.(2016)Huang, Sun, Liu, Sedra, and
  Weinberger}]{stochdepth}
Gao Huang, Yu~Sun, Zhuang Liu, Daniel Sedra, and Kilian~Q. Weinberger. 2016.
\newblock \href {https://arxiv.org/abs/1603.09382} {{Deep Networks with
  Stochastic Depth}}.
\newblock In \emph{Computer Vision -- ECCV 2016}.

\bibitem[{Jeddi et~al.(2026)Jeddi, Ciccone, and Taati}]{loopformer}
Ahmadreza Jeddi, Marco Ciccone, and Babak Taati. 2026.
\newblock \href {https://arxiv.org/abs/2602.11451} {{LoopFormer: Elastic-Depth
  Looped Transformers for Latent Reasoning via Shortcut Modulation}}.
\newblock \emph{arXiv preprint arXiv:2602.11451}.

\bibitem[{Liang et~al.(2026)Liang, Zhang, Du, Wu, Ma, Pu, Zhang, Wang, and
  Lin}]{asymloop}
Cheng Liang, Tengfei Zhang, Wenjie Du, Jianyu Wu, Changlian Ma, Shuchen Pu,
  Ya~Zhang, Yanfeng Wang, and Zhouhan Lin. 2026.
\newblock \href {https://arxiv.org/abs/2610.28489} {{Asymmetric Looped
  Transformers: Fixed Global Representations for an Iterative Decoder}}.
\newblock \emph{arXiv preprint arXiv:2610.28489}.

\bibitem[{Merity et~al.(2017)Merity, Xiong, Bradbury, and Socher}]{wikitext}
Stephen Merity, Caiming Xiong, James Bradbury, and Richard Socher. 2017.
\newblock \href {https://arxiv.org/abs/1609.07843} {{Pointer Sentinel Mixture
  Models}}.
\newblock In \emph{International Conference on Learning Representations}.

\bibitem[{O'Neill and Reid(2026)}]{lla}
James O'Neill and Fergal Reid. 2026.
\newblock \href {https://arxiv.org/abs/2607.15456} {{Looped Latent Attention:
  Cross-Loop KV Compression for Looped Transformers}}.
\newblock \emph{arXiv preprint arXiv:2607.15456}.

\bibitem[{Paperno et~al.(2016)Paperno, Kruszewski, Lazaridou, Pham, Bernardi,
  Pezzelle, Baroni, Boleda, and Fern{\'a}ndez}]{lambada}
Denis Paperno, Germ{\'a}n Kruszewski, Angeliki Lazaridou, Ngoc~Quan Pham,
  Raffaella Bernardi, Sandro Pezzelle, Marco Baroni, Gemma Boleda, and Raquel
  Fern{\'a}ndez. 2016.
\newblock \href {https://aclanthology.org/P16-1144/} {{The LAMBADA Dataset:
  Word Prediction Requiring a Broad Discourse Context}}.
\newblock In \emph{Proceedings of the 54th Annual Meeting of the Association
  for Computational Linguistics}.

\bibitem[{Penedo et~al.(2024)Penedo, Kydl{\'i}{\v c}ek, Ben~Allal, Lozhkov,
  Mitchell, Raffel, von Werra, and Wolf}]{finewebedu}
Guilherme Penedo, Hynek Kydl{\'i}{\v c}ek, Loubna Ben~Allal, Anton Lozhkov,
  Margaret Mitchell, Colin Raffel, Leandro von Werra, and Thomas Wolf. 2024.
\newblock \href {https://arxiv.org/abs/2406.17557} {{The FineWeb Datasets:
  Decanting the Web for the Finest Text Data at Scale}}.
\newblock In \emph{Advances in Neural Information Processing Systems, Datasets
  and Benchmarks Track}.

\bibitem[{Peng et~al.(2024)Peng, Quesnelle, Fan, and Shippole}]{yarn}
Bowen Peng, Jeffrey Quesnelle, Honglu Fan, and Enrico Shippole. 2024.
\newblock \href {https://arxiv.org/abs/2309.00071} {{YaRN: Efficient Context
  Window Extension of Large Language Models}}.
\newblock In \emph{International Conference on Learning Representations}.

\bibitem[{Prairie et~al.(2026)Prairie, Ryan, Berg-Kirkpatrick, and Fu}]{benice}
Hayden Prairie, Jacob~M. Ryan, Taylor Berg-Kirkpatrick, and Daniel~Y. Fu. 2026.
\newblock \href {https://arxiv.org/abs/2610.31633} {{BeNice! Closing Looping's
  Inference-Memory Gap via KV-Sharing}}.
\newblock \emph{arXiv preprint arXiv:2610.31633}.

\bibitem[{Rae et~al.(2020)Rae, Potapenko, Jayakumar, and Lillicrap}]{pg19}
Jack~W. Rae, Anna Potapenko, Siddhant~M. Jayakumar, and Timothy~P. Lillicrap.
  2020.
\newblock \href {https://arxiv.org/abs/1911.05507} {{Compressive Transformers
  for Long-Range Sequence Modelling}}.
\newblock In \emph{International Conference on Learning Representations}.

\bibitem[{Raposo et~al.(2024)Raposo, Ritter, Richards, Lillicrap, Humphreys,
  and Santoro}]{mod}
David Raposo, Sam Ritter, Blake Richards, Timothy Lillicrap, Peter~Conway
  Humphreys, and Adam Santoro. 2024.
\newblock \href {https://arxiv.org/abs/2404.02258} {{Mixture-of-Depths:
  Dynamically allocating compute in transformer-based language models}}.
\newblock \emph{arXiv preprint arXiv:2404.02258}.

\bibitem[{Sakaguchi et~al.(2020)Sakaguchi, Le~Bras, Bhagavatula, and
  Choi}]{winogrande}
Keisuke Sakaguchi, Ronan Le~Bras, Chandra Bhagavatula, and Yejin Choi. 2020.
\newblock \href {https://ojs.aaai.org/index.php/AAAI/article/view/6399}
  {{WinoGrande: An Adversarial Winograd Schema Challenge at Scale}}.
\newblock In \emph{Proceedings of the AAAI Conference on Artificial
  Intelligence}.

\bibitem[{Schuster et~al.(2022)Schuster, Fisch, Gupta, Dehghani, Bahri, Tran,
  Tay, and Metzler}]{calm}
Tal Schuster, Adam Fisch, Jai Gupta, Mostafa Dehghani, Dara Bahri, Vinh Tran,
  Yi~Tay, and Donald Metzler. 2022.
\newblock \href {https://arxiv.org/abs/2207.07061} {{Confident Adaptive
  Language Modeling}}.
\newblock In \emph{Advances in Neural Information Processing Systems}.

\bibitem[{Sun et~al.(2024)Sun, Dong, Zhu, Huang, Wang, Ma, Zhang, Wang, and
  Wei}]{yoco}
Yutao Sun, Li~Dong, Yi~Zhu, Shaohan Huang, Wenhui Wang, Shuming Ma, Quanlu
  Zhang, Jianyong Wang, and Furu Wei. 2024.
\newblock \href {https://arxiv.org/abs/2405.05254} {{You Only Cache Once:
  Decoder-Decoder Architectures for Language Models}}.
\newblock In \emph{Advances in Neural Information Processing Systems}.

\bibitem[{Wu and Tu(2024)}]{lckv}
Haoyi Wu and Kewei Tu. 2024.
\newblock \href {https://doi.org/10.18653/v1/2024.acl-long.602}
  {{Layer-Condensed KV Cache for Efficient Inference of Large Language
  Models}}.
\newblock In \emph{Proceedings of the 62nd Annual Meeting of the Association
  for Computational Linguistics (Volume 1: Long Papers)}, pages 11175--11188.

\bibitem[{Xie et~al.(2025)Xie, Wei, Cao, Zhao, Deng, Li, Dai, Gao, Chang, Yu,
  Zhao, Zhou, Xu, Zhang, Zeng, Hu, Wang, Yuan, Wang, and Liang}]{mhc}
Zhenda Xie, Yixuan Wei, Huanqi Cao, Chenggang Zhao, Chengqi Deng, Jiashi Li,
  Damai Dai, Huazuo Gao, Jiang Chang, Kuai Yu, Liang Zhao, Shangyan Zhou, Zhean
  Xu, Zhengyan Zhang, Wangding Zeng, Shengding Hu, Yuqing Wang, Jingyang Yuan,
  Lean Wang, and Wenfeng Liang. 2025.
\newblock \href {https://arxiv.org/abs/2512.24880} {{mHC: Manifold-Constrained
  Hyper-Connections}}.
\newblock \emph{arXiv preprint arXiv:2512.24880}.

\bibitem[{Yu and Huang(2019)}]{usnet}
Jiahui Yu and Thomas~S. Huang. 2019.
\newblock \href {https://arxiv.org/abs/1903.05134} {{Universally Slimmable
  Networks and Improved Training Techniques}}.
\newblock In \emph{Proceedings of the IEEE/CVF International Conference on
  Computer Vision}.

\bibitem[{Zellers et~al.(2019)Zellers, Holtzman, Bisk, Farhadi, and
  Choi}]{hellaswag}
Rowan Zellers, Ari Holtzman, Yonatan Bisk, Ali Farhadi, and Yejin Choi. 2019.
\newblock \href {https://aclanthology.org/P19-1472/} {{HellaSwag: Can a Machine
  Really Finish Your Sentence?}}
\newblock In \emph{Proceedings of the 57th Annual Meeting of the Association
  for Computational Linguistics}.

\bibitem[{Zhou et~al.(2026)Zhou, Yu, Weng, Zhou, Yin, Yang, and Pan}]{buddy}
Yuhua Zhou, Shaoqi Yu, Shichao Weng, Changhai Zhou, Mingze Yin, Fei Yang, and
  Aimin Pan. 2026.
\newblock \href {https://arxiv.org/abs/2606.09514} {{BUDDY: BUdget-Driven
  DYnamic Depth Routing for Adaptive Large Language Model Inference}}.
\newblock \emph{arXiv preprint arXiv:2606.09514}.

\bibitem[{Zhu et~al.(2025{\natexlab{a}})Zhu, Huang, Huang, Zeng, Mao, Wu, Min,
  and Zhou}]{hyperconnections}
Defa Zhu, Hongzhi Huang, Zihao Huang, Yutao Zeng, Yunyao Mao, Banggu Wu, Qiyang
  Min, and Xun Zhou. 2025{\natexlab{a}}.
\newblock \href {https://arxiv.org/abs/2409.19606} {{Hyper-Connections}}.
\newblock In \emph{International Conference on Learning Representations}.

\bibitem[{Zhu et~al.(2025{\natexlab{b}})Zhu, Wang, Hua, Zhang, Li, Que, Wei,
  Wen, Yin, Xing, Li, Shi, Ma, Li, Kergan, Smith, Qu, Hui, Wu, Min, Huang,
  Zhou, Ye, Liu, Yang, Shi, Lin, Zhao, Cai, Zhang, Huang, Bengio, and
  Eshraghian}]{ouro}
Rui-Jie Zhu, Zixuan Wang, Kai Hua, Tianyu Zhang, Ziniu Li, Haoran Que, Boyi
  Wei, Zixin Wen, Fan Yin, He~Xing, Lu~Li, Jiajun Shi, Kaijing Ma, Shanda Li,
  Taylor Kergan, Andrew Smith, Xingwei Qu, Mude Hui, Bohong Wu, and 14 others.
  2025{\natexlab{b}}.
\newblock \href {https://arxiv.org/abs/2510.25741} {{Scaling Latent Reasoning
  via Looped Language Models}}.
\newblock \emph{arXiv preprint arXiv:2510.25741}.

\end{thebibliography}
\clearpage
\appendix
\section{Architecture and Training Details}
\label{app:training}
\begin{table}[t]
\centering\small
\setlength{\tabcolsep}{4pt}
\begin{tabular}{lr}
\toprule
Property & Value \\
\midrule
Backbone parameters & 204,046,232 \\
Planner parameters (planner model only) & 19,159 \\
Hidden / feed-forward width & 832 / 2,688 \\
Attention heads $\times$ head dimension & 64 $\times$ 80 \\
Sliding-window width (enc.\ 1--2; dec.) & 256; 128 \\
Indexer heads $\times$ width, Top-K & 32 $\times$ 32, 128 \\
Global KV compression (enc.\ 3/5; dec.\ 1) & 2; 1 \\
Reader layers & 5 \\
Decoder calls per token & 3--21 \\
Training targets & 5,000,200,876 \\
\bottomrule
\end{tabular}
\caption{Architecture of the shared backbone. KV: key--value.}
\label{tab:architecture}
\end{table}

\paragraph{Optimization.}
We use AdamW with a peak learning rate of $3\times10^{-4}$ that decays to $3\times10^{-5}$, 114 warmup 
updates, a weight decay of 0.1, and gradient clipping at a maximum norm of 1.0. Computation uses BF16, with 
selected components kept in FP32.

\paragraph{Attention and residual structure.}
The backbone uses CSA2 compressed-sparse attention \citep{ced}, whose three modes are defined in 
Section~\ref{sec:backbone}. Inside the encoder, layers 3 and 5 write compressed KV at ratio two, which layers 
4 and 6 reuse. Decoder layers 2, 4 and 6 reuse the latest decoder retrieval support. Query low-rank width is 
208, with eight output groups and an output low-rank width of 160. The residual structure uses four parallel
streams of manifold-constrained hyper-connections \citep{mhc,hyperconnections}, and the input embeddings and
the output head are untied. Global KV producers are encoder layers 3 and 5 and decoder layer 1 (physical layers
L3, L5 and L7), with compression ratios two, two and one, respectively. Index producers are encoder layers 3 
and 5 and decoder layers 1, 3 and 5. Decoder layers 3 and 5 recompute queries and Top-128 support over 
decoder layer 1's global bank but do not append persistent global KV or indexer keys. Decoder layers 4 and 6 
consume the selections from their most recent visit. This reuse violates neither the pre-repetition 
global-memory rule nor the cache size of Eq.~\eqref{eq:cachebound}.

\paragraph{Planner.}
The planner reads $u_t\in\R^{832}$, the normalized output of decoder layer 1. The score groups in 
Eq.~\eqref{eq:plannerscore} have sizes $(5,4,5,4,5)$. Each reader computes candidate hidden states for depths
of up to four.

In Eq.~\eqref{eq:plannerloss}, $\mathcal H$ is the entropy of a token's routing distribution, averaged over 
valid tokens and variable readers.

Under a per-token cap $B$, the allocator solves $\max_{d:\,C(d)\le B}\sum_\ell z_{t,\ell,d_\ell}$ with a 
multiple-choice dynamic program in $O(L_D B\max_j|\mathcal D_j|)$ time per token. Because the cap is an upper 
bound, realized calls can be lower. Replacing the max with a log-sum-exp gives the constrained marginals used 
by the budget-conditioned continuation (Appendix~\ref{app:budget}); that continuation adds a zero-initialized 
budget table $e_B\in\R^{23}$ with 22 rows and 506 parameters.

\paragraph{Scalar-depth planner.}\phantomsection\label{app:scalar}%
A coarser planner predicts one depth per token and shares it across decoder layers 2--6, so it cannot skip 
a layer. It reaches a test CE of 3.04026 at 19.17 calls on average, behind the layerwise planner (3.03546 at 
17.96 native calls) and Random-Depth at 19 calls (3.03058).

\paragraph{Dense execution and sampling.}
Each microbatch is unrolled to its
largest sampled depth, unexecuted repetitions are masked, and each KV entry is still published only once.
Sampling is deterministic and seeded by the training seed together with the update, microbatch slot and
data-parallel rank.

Self-Distilled draws one budget per microbatch slot and a path per rank, with the encoder and decoder layer 1
forming a prefix shared by student and full-depth teacher.

\paragraph{Static path search.}
The staged search of Section~\ref{sec:static} is exhaustive only at its first stage, so it does not certify 
the best static path on all 512 samples. All selection files record that paths were locked before test 
scoring.

\section{Uniform-Budget Control}
\label{app:uniform}
\begin{table}[t]
\centering\small
\setlength{\tabcolsep}{4pt}
\begin{tabular}{rrr}
\toprule
Calls & Deep-heavy & Uniform \\
\midrule
\RowsRecipeEarly
\midrule
Train calls & 17.19 & 13.88 \\
\bottomrule
\end{tabular}
\caption{Two budget distributions in the same recipe, compared early in training: test cross-entropy (CE;
nats per token, lower is better). Deep-heavy is Random-Depth's schedule; the uniform run instead draws the
full 21-call path with probability 0.25 and otherwise a budget uniform in 3--20 calls. Both then draw a
uniformly random legal path of that budget, skipped layers included. Train calls are expected decoder calls
per training token; the uniform run uses fewer, which confounds absolute levels but not the within-model
flatness of its curve.}
\label{tab:recipes}
\end{table}
Section~\ref{sec:recipes} attributes Random-Depth's weak low budgets to how rarely its deep-heavy schedule
samples them, rather than to shallow execution itself. Self-Distilled supports this but also adds a full-depth
teacher, so it changes two things at once. The uniform run changes only the budget distribution: it keeps
Random-Depth's recipe but spreads most samples evenly over 3--20 calls instead of placing three quarters at
17--21. Both runs are compared at the same early point in training.

Table~\ref{tab:recipes} gives the schedule and the outcome. The uniform run is better at 3 and 6 calls, so
shallow paths become strong once they are trained often enough, and the deep-heavy schedule is better from 11
calls on; Section~\ref{sec:recipes} reads what that trade costs.

\section{Cost and Prefill Measurements}
\label{app:cost}
\paragraph{Protocol.}
Decode uses a 32-token teacher-forced continuation after a 128-token prompt, with eight requests and three 
repetitions per path. The Python reference loop times the encoder, selected decoder bodies, cache updates and 
FP32 logits. Matrix FLOPs are counted as in Section~\ref{sec:setup}, from dispatched operand shapes including 
explicit tile padding and masked bank rows; they are neither hardware instruction counts nor theoretical 
FLOPs for useful tokens. Each row averages those requests whose operator-count audit passed, between six and 
eight of the eight. Cache size is measured as live persistent storage. Fixed4 was not timed independently at 
21 calls; we reuse the identical path shape, and the two measured models differ by at most 
\LatencyCheckpointGapPct\% on identical paths.

\begin{figure*}[t]
\centering
\includegraphics[width=\textwidth]{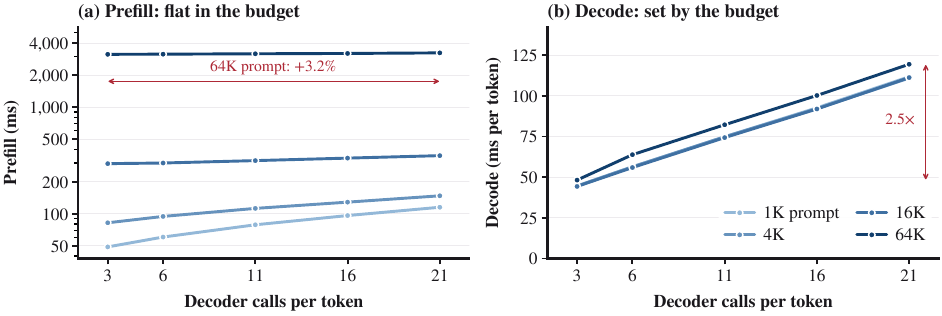}
\caption{Prefill and subsequent decode for Random-Depth against the budget, one curve per prompt
length (one \TimingGPU{}, batch size one, median of five). (a) Prefill is set by the prompt, not the
budget: at 65,536 tokens the whole 3-to-21-call range costs \PrefillLongIncreasePct\% more. (b) Decode
is set by the budget, with the four prompt lengths almost on one curve. The persistent cache is
identical across paths at every prompt length. Prompts beyond 4,096 tokens are timed for cost only.
\PrefillRemeasuredNote}
\label{fig:prefill}
\end{figure*}
\paragraph{Decode and prefill.}
On identical static paths, cost depends on the path and model shape rather than on the weights. With 512- and 
2,048-token prompts, the persistent cache is again identical on the 3- and 21-call paths. Native routing 
takes \LatencyNativeMs\,ms at \LatencyNativeCalls{} calls, within \LatencyNativeInterpPct\% of interpolation 
between the static 16- and 21-call paths. Prefill follows DeepSeek-V4.1-Flash's SWA replay of the most recent 
128 prompt tokens \citep{ced}. At 65,536 prompt tokens, prefill takes \PrefillLongThreeS\,s at 3 calls and 
\PrefillLongTwentyOneS\,s at 21, an increase of \PrefillLongIncreasePct\%. The budget therefore mainly 
affects decode (Figure~\ref{fig:prefill}).

\section{Grouped Decoder Memory}
\label{app:grouped}
The comparison uses Fixed4's schedule: encoder layers 1--6 and decoder layer 1 once, then decoder layers 2--6 
four times each. LooM's layout builds one decoder memory at decoder layer 1 and reuses it through decoder 
layer 6. The grouped layout builds memories at decoder layers 1, 3 and 5, adding two global producers and 
138,464 parameters; later memories are computed after the repeated reads. The grouped model increases 
persistent cache size by 1.58$\times$ at 1,024 prompt tokens, 1.80$\times$ at 4,096 and 1.88$\times$ at 
16,384. The paired interval is computed over the same 7,836 test document segments with 2,000 resamples. This 
is one Fixed4-schedule variant and one seed, not a general test of all memory-sharing designs.

\section{Recirculating the Whole Decoder}
\label{app:ouro}
The whole-decoder control repeats the six-layer decoder up to four times with shared weights, with each round 
retaining its own local history. The encoder and global memory are computed once. A gate defines a first-exit 
distribution, and training minimizes expected CE minus an entropy bonus with coefficient 0.1. The run starts 
from a random initialization and reuses the single-pass data, optimizer, schedule and seed.

At update 1,800, on the same 512-sample, 1,737,712-target validation set, the whole-decoder loop reaches 3.3814, 3.3557, 3.3538 and 3.3541 over one to four rounds. 
The second round reduces CE by 0.026 nats, the third by 0.002, and the fourth yields no further reduction, so
no fixed number of rounds reaches the single-pass model at 3.3496. The learned exit distribution is close to
uniform, $(0.21,0.27,0.26,0.26)$. Random-Depth at 21 calls reaches 3.3458 on this snapshot, but the two objectives differ
and the run is a single-seed preliminary study, so we cannot attribute that difference to reading fixed 
memory.

\section{Planner Continuation and Allocation Diagnostics}
\label{app:budget}
\begin{table}[t]
\centering\footnotesize
\setlength{\tabcolsep}{4pt}
\begin{tabular}{lrl}
\toprule
Replacement & Cap & $\Delta$CE \\
\midrule
\RowsAllocDiag
\bottomrule
\end{tabular}
\caption{Allocation controls on two 200-update budget-conditioned planner continuations (seeds 42 and 43), needed
because the original planner is poorly calibrated at cap 6. Each value is the range of replacement minus
native-routing cross-entropy (CE); negative favors the replacement. At cap 6, static path and static split improve over poorly calibrated native routing. A cap bounds each token's total calls.}
\label{tab:allocdiagmain}
\end{table}

The original planner is poorly calibrated at small caps, so we continued it and its budget table for 200 
updates under caps $B\in\{6,11,14,16,21\}$, using a zero-initialized budget embedding. The two continuations 
(seeds 42 and 43) are used only for the allocation controls; the main frontier compares the original 
5,703-update checkpoints. Their native-CE differences across caps span \ContinuationSeedRange{} nats. These 
are continuation-seed differences, not independent pretraining replicates.

Random-Depth and the planner differ in more than token dependence: their training also differs in auxiliary
losses, exploration noise and straight-through gradients. At 16 and 21 calls, Random-Depth's margins are 0.0067 and 
0.0049 nats, too small to establish a seed-robust ordering. Native routing costs what its mean call count predicts: it falls 
within \LatencyNativeInterpPct\% of interpolation between the static paths (Appendix~\ref{app:cost}). A 
scalar planner and longer continuations improve low budgets, but neither rules out a usable token-level 
signal (Appendix~\ref{app:scalar}).

\section{Mechanism Analysis: Protocol and Gains by Decile}
\label{app:mechanism}
\begin{figure*}[t]
\centering
\includegraphics[width=\textwidth]{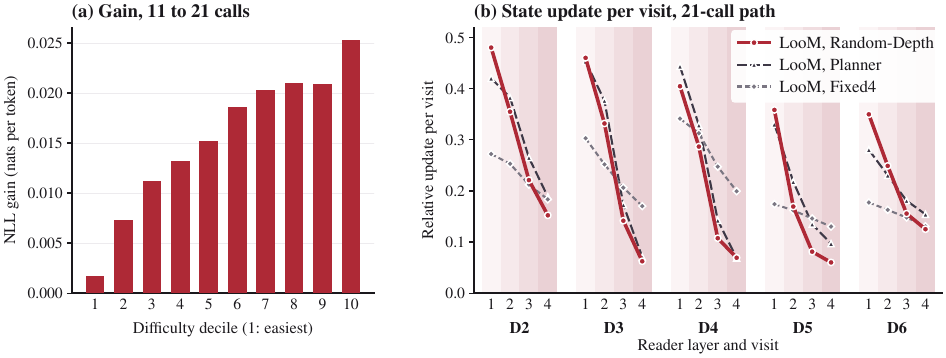}
\caption{Mechanism analysis on 256 test sequences. (a) The allocation-relevant negative log-likelihood
(NLL) gain from 11 to 21 calls by difficulty decile (1: easiest); positive is improvement, and difficulty
uses mean NLL over both paths, avoiding selection on one arm's noise. (b) Relative residual-state updates
at visits 1--4 of decoder layers 2--6, labeled D2--D6 in the panel, on the 21-call path.}
\label{fig:mechanism}
\end{figure*}
Figure~\ref{fig:mechanism} reports the full analysis. Random-Depth is scored on locked 11- and 21-call
paths, whereas Fixed4 and the planner use 21 calls, all on the same 256 test sequences (865,148 targets). 
State and retrieval statistics use the first 64 sequences. Difficulty deciles use the mean NLL of the
compared paths, which avoids binning on one arm's noise. In the retrieval-freezing intervention,
visits 2--4 of decoder layers 3 and 5 reuse their first-visit Top-$K$ sets, so decoder layers 4 and 6 also
read those sets. Scores are still computed; only the selection is replaced.

\paragraph{Relative state update.}
Panel~(b) measures how far one visit moves a token's hidden state. For a visit that takes the state from
$h_{\mathrm{in}}$ to $h_{\mathrm{out}}$, the relative update is
$\lVert h_{\mathrm{out}}-h_{\mathrm{in}}\rVert/\lVert h_{\mathrm{in}}\rVert$, averaged over tokens.
Dividing by the incoming norm is what makes the five reader layers comparable, since their states differ
in scale: a value of 0.48 means the visit moved the state by about half its own length, and 0.15 means it
moved it by a seventh. The quantity is a distance and carries no direction, so it shows that later visits
keep changing the state, not that the change helps. What ties repeated reading to quality is the
retrieval-freezing intervention of Section~\ref{sec:reads}.

\section{Downstream Evaluation}
\label{app:downstream}
\paragraph{Protocol and configurations.}
We evaluate with lm-evaluation-harness 0.4.13 \citep{lmeval} on LAMBADA \citep{lambada}, HellaSwag 
\citep{hellaswag}, PIQA \citep{piqa}, WinoGrande \citep{winogrande}, ARC-Easy and ARC-Challenge \citep{arc}, 
reporting accuracy, or normalized accuracy for HellaSwag and ARC-Challenge. We fixed the metrics and macro 
average before scoring. The full sets contain 5,153, 10,042, 1,838, 1,267, 2,376 and 1,172 examples, 
respectively. WikiText-2 uses 62 documents with rolling 4,096-token windows \citep{wikitext}. We evaluated 24 
configurations: single-pass, Fixed4, native routing, and the 3-, 6-, 8-, 11-, 13-, 16- and 21-call static 
paths of Random-Depth, Self-Distilled and the planner. Static configurations use the paths locked for the CE 
frontier.

\paragraph{Comparisons with intervals crossing zero.}
Random-Depth at 11 calls is 0.71 points above Fixed4 at 21 on the six-task average, but the interval 
$[-0.04,+1.43]$ crosses zero. Self-Distilled versus Random-Depth at six calls is likewise unresolved 
($+0.40$, $[-0.38,+1.17]$). WikiText-2 requires 16 rather than 11 calls to overtake Fixed4 at 21, so the 
11-call CE equality is distribution-specific. Because scores are non-monotonic across 13, 16 and 21 calls, 
sub-point differences in the aggregate should not be interpreted as a frontier ordering.

\section{Long-Context Continuation and Retrieval Protocol}
\label{app:long}
Random-Depth, the single-pass model and the planner start from their 5B-target checkpoints and receive the 
same 600-update continuation (314,572,800 targets per model). The continuation consists of 8K, 16K, 32K and 
64K phases with 60, 60, 120 and 360 updates, respectively. At every phase, two thirds of the updates use long 
PG19 examples and one third use short-domain data. All three models use the same data order, optimizer and 
learning-rate schedule ($3\times10^{-5}$ to $3\times10^{-6}$, with 60 warmup updates). The 45 PG19 books used 
for evaluation are disjoint from the continuation's training and validation books. The planner loads its 
trained routing parameters and continues straight-through training with a balance coefficient of 0.001, no 
entropy bonus and no exploration noise; an audit verifies that \texttt{planner.weight} changes. Random-Depth
instead keeps its pretraining schedule (17.19 expected calls). The compressed-attention
branch uses YaRN/NTK-by-parts-derived interpolation targeting 64K \citep{yarn}, whereas the local branch is 
unchanged.

Perplexity is evaluated before task fine-tuning. Random-Depth minus planner static-21 is $+0.0031$ nats at 
4K ($[-0.0002,+0.0063]$) and $+0.0050$ at 64K ($[+0.0014,+0.0084]$). Planner static-21 minus native routing 
ranges from $+0.0008$ to $+0.0014$ nats, with every interval excluding zero. Table~\ref{tab:longmain} gives 
the compact Random-Depth--single-pass comparison; Table~\ref{tab:longplannerdiff} reports the paired routing 
differences with whole-book intervals.

\begin{table}[t]
\centering\small
\setlength{\tabcolsep}{4pt}
\begin{tabular}{lll}
\toprule
Ctx. & RD $-$ native & Static $-$ native \\
\midrule
\RowsLongPlannerDiff
\bottomrule
\end{tabular}
\caption{Paired negative log-likelihood (NLL) differences in $10^{-3}$ nats per token with whole-book 95\% confidence
intervals; negative favors the first model in each difference. RD denotes Random-Depth; Static and native
denote the planner with its static path and with native routing, respectively. Point estimates favor native at
every length: static minus native excludes zero at every length, while RD minus native excludes zero only at
4K and 64K.}
\label{tab:longplannerdiff}
\end{table}

For retrieval, all continued models receive the same direct-answer fine-tuning on 512 examples (64 updates, 
response-only supervision, and no 64K examples). Random-Depth draws sampled depths, the single-pass model 
runs one pass, and the planner uses native or static-21 execution. We evaluate 
\texttt{niah\_single\_1}, \texttt{niah\_single\_2} and \texttt{niah\_multikey\_1} from RULER \citep{ruler} at 
4K, 8K, 16K, 32K and 64K, with 20 questions per task and context length. This is a 300-question subset, not 
the full 13-task RULER benchmark.

\begin{table}[t]
\centering\footnotesize
\setlength{\tabcolsep}{4pt}
\begin{tabular}{lrrrr}
\toprule
Length & Single & Planner native & Planner static & RD \\
\midrule
\RowsLongNiah
\bottomrule
\end{tabular}
\caption{Three-task needle-in-a-haystack (NIAH) macro accuracy (\%, higher is better) by context length in tokens and
overall (300 questions per model). Single denotes the single-pass model, and RD denotes Random-Depth. All arms
use the same answer-format fine-tuning. Random-Depth and the planner with its static path use the static
21-call path, whereas the planner with native routing uses its learned per-token policy. The table reports
point estimates; the overall paired interval for RD minus the single-pass model is $[+1.33,+10.33]$ percentage
points.}
\label{tab:niah}
\end{table}

\section{Executed Hindsight Allocation}
\label{app:oracle}

We use the same 256 test sequences and 865,148 targets as in the mechanism analysis. Candidate paths are 
Random-Depth's validation-locked 11-, 13-, 16- and 21-call paths. For each token, a Lagrangian hindsight 
policy uses the candidate NLLs to select a path under an average budget of 16 calls. A greedy assignment 
followed by a dynamic-programming repair enforces that total exactly. The repair is not a certified global 
optimum, so the measured gain is a lower bound on the best hindsight gain over these four paths.

The mixed execution reaches CE \OracleCE{}, compared with \OracleStaticCE{} for static 16, a gain of
\OracleGain{} nats. Scoring each token on its chosen path without executing the mixture reaches 2.9514, so
leaving out the execution effects overstates the gain by 0.0117 nats. This is a label-aware upper-bound
reference, not evidence of prefix-predictable signal or
of a deployable allocation policy.

\section{Data}
\label{app:data}
Training draws 5B valid target tokens from the FineWeb-Edu 100BT sample \citep{finewebedu}, tokenized with
the SmolLM2 tokenizer \citep{smollm2}, whose vocabulary holds 49,152 tokens. Deterministic packed manifests
record source-document segments and target masks. Source documents are hashed, and each one is assigned
whole to exactly one of the training, validation and test partitions, so no document contributes to more
than one of them; the packed samples are then shuffled once.

\end{document}